\documentclass[11pt]{article}
\pdfoutput=1
\usepackage{latexsym,amsmath,amssymb,graphicx}
\usepackage{hyperref}
\usepackage[title]{appendix}

\advance \topmargin by -\headheight
\advance \topmargin by -\headsep
\evensidemargin \oddsidemargin
\title{Accelerating Self-Play Learning in Go}
\author{David J. Wu\thanks{email:lightvector@gmail.com. Many thanks to Craig Falls, David Parkes, James Somers, and numerous others for their kind feedback and advice on this work.} \\ Jane Street Group}
\begin{document}
\maketitle
\begin{abstract}
By introducing several improvements to the AlphaZero process and architecture, we greatly accelerate self-play learning in Go, achieving a 50x reduction in computation over comparable methods. Like AlphaZero and replications such as ELF OpenGo and Leela Zero, our bot KataGo only learns from neural-net-guided Monte Carlo tree search self-play. But whereas AlphaZero required thousands of TPUs over several days and ELF required thousands of GPUs over two weeks, KataGo surpasses ELF's final model after only 19 days on fewer than 30 GPUs. Much of the speedup involves non-domain-specific improvements that might directly transfer to other problems. Further gains from domain-specific techniques reveal the remaining efficiency gap between the best methods and purely general methods such as AlphaZero. Our work is a step towards making learning in state spaces as large as Go possible without large-scale computational resources.
\end{abstract}

\section{Introduction}
In 2017, DeepMind's AlphaGoZero demonstrated that it was possible to achieve superhuman performance in Go without reliance on human strategic knowledge or preexisting data \cite{AGZ}. Subsequently, DeepMind's AlphaZero achieved comparable results in Chess and Shogi. However, the amount of computation required was large, with DeepMind's main reported run for Go using 5000 TPUs for several days, totaling about 41 TPU-years \cite{AZ}. Similarly ELF OpenGo, a replication by Facebook, used 2000 V100 GPUs for about 13-14 days\footnotemark, or about 74 GPU-years, to reach top levels of performance \cite{FB2}.

\addtocounter{footnote}{-1}
\stepcounter{footnote}\footnotetext{Although ELF's training run lasted about 16 days, its final model was chosen from a point slightly prior to the end of the run.}

In this paper, we introduce several new techniques to improve the efficiency of self-play learning, while also reviving some pre-AlphaZero ideas in computer Go and newly applying them to the AlphaZero process. Although our bot KataGo uses some domain-specific features and optimizations, it still starts from random play and makes no use of outside strategic knowledge or preexisting data. It surpasses the strength of ELF OpenGo after training on about 27 V100 GPUs for 19 days, a total of about 1.4 GPU-years, or about a factor of 50 reduction. And by a conservative comparison, KataGo is also at least an order of magnitude more efficient than the multi-year-long online distributed training project Leela Zero \cite{LeelaZero}. Our code is open-source, and superhuman trained models and data from our main run are available online\footnotemark.

\footnotetext{https://github.com/lightvector/KataGo. Using our code, it is possible to reach strong or top human amateur strength starting from nothing on even single GPUs in mere days, and several people have in fact already done so.}

We make two main contributions:

First, we present a variety of domain-independent improvements that might directly transfer to other AlphaZero-like learning or to reinforcement learning more generally. These include: (1) a new technique of \emph{playout cap randomization} to improve the balance of data for different targets in the AlphaZero process, (2) a new technique of \emph{policy target pruning} that improves policy training by decoupling it from exploration in MCTS, (3) the addition of a \emph{global-pooling} mechanism to the neural net, agreeing with research elsewhere on global context in image tasks such as by Hu et al. (2018) \cite{SE}, and (4) a revived idea from supervised learning in Go to add \emph{auxiliary policy targets} from future actions tried by Tian and Zhu (2016) \cite{Darkforest}, which we find transfers easily to self-play and could apply widely to other problems in reinforcement learning.

Second, our work serves as a case study that there is still a significant efficiency gap between AlphaZero's methods and what is possible from self-play. We find nontrivial further gains from some domain-specific methods. These include \emph{auxiliary ownership and score targets} (similar to those in Wu et al. 2018 \cite{MLVGo}) and which actually also suggest a much more general meta-learning heuristic: that predicting subcomponents of desired targets can greatly improve training. We also find that adding some game-specific input features still significantly improves learning, indicating that though AlphaZero succeeds without them, it is also far from obsoleting them. 

In Section \ref{Overview} we summarize our architecture. In Sections \ref{GeneralSection} and \ref{SpecificSection} we outline the general techniques of \emph{playout cap randomization}, \emph{policy target pruning}, \emph{global-pooling}, and \emph{auxiliary policy targets}, followed by domain-specific improvements including \emph{ownership and score targets} and input features. In Section \ref{Experiments} we present our data, including comparison runs showing how these techniques each improve learning and all similarly contribute to the final result.


\section{Basic Architecture and Parameters} \label{Overview}

Although varying in many minor details, KataGo's overall architecture resembles the AlphaGoZero and AlphaZero architectures \cite{AGZ,AZ}.

\enlargethispage{-\baselineskip}
KataGo plays games against itself using Monte-Carlo tree search (MCTS) guided by a neural net to generate training data. Search consists of growing a game tree by repeated playouts. Playouts start from the root and descend the tree, at each node $n$ choosing the child $c$ that maximizes:
\[ \text{PUCT}(c) = V(c) + c_{\text{PUCT}} P(c) \frac{ \sqrt{\sum_{c'} N(c')} } { 1 + N(c) } \]
where $V(c)$ is the average predicted utility of all nodes in $c$'s subtree, $P(c)$ is the policy prior of $c$ from the neural net, $N(c)$ is the number of playouts previously sent through child $c$, and $c_{\text{PUCT}} = 1.1$. Upon reaching the end of the tree and finding that the next chosen child is not allocated, the playout terminates by appending that single child to the tree.\footnotemark

\footnotetext{
When $N(c) = 0$ and $V(c)$ is undefined, unlike AlphaZero but like Leela Zero, we define:
$ V(c) = V(n) - c_{\text{FPU}} \sqrt{ P_{\text{explored}} } $
where $P_{\text{explored}} = \sum_{c' | N(c') > 0} P(c')$ is the total policy of explored children and $c_{\text{FPU}} = 0.2$ is a ``first-play-urgency'' reduction coefficient, except $c_{\text{FPU}} = 0$ at the root if Dirichlet noise is enabled. 
}

Like AlphaZero, to aid discovery of unexpected moves, KataGo adds noise to the policy prior at the root: 
\[ P(c) = 0.75 P_{\text{raw}}(c) + 0.25 \, \eta \]
where $\eta$ is a draw from a Dirichlet distribution on legal moves with parameter $\alpha = 0.03 * 19^2/\text{N}(c)$ where $N$ is the total number of legal moves. This matches AlphaZero's $\alpha = 0.03$ on the empty $19 \times 19$ Go board while scaling to other sizes. KataGo also applies a softmax temperature at the root of $1.03$, an idea to improve policy convergence stability from SAI, another AlphaGoZero replication \cite{SAI} .

The neural net guiding search is a convolutional residual net with a \emph{preactivation} architecture \cite{IDMapRes}, with a trunk of $b$ residual blocks with $c$ channels. Similar to Leela Zero \cite{LeelaZero}, KataGo began with small nets and progressively increased their size, concurrently training the next larger size on the same data and switching when its average loss caught up to the smaller size. In KataGo's main 19-day run, $(b,c)$ began at $(6,96)$ and switched to $(10,128)$, $(15,192)$, and $(20,256)$, at roughly 0.75 days, 1.75 days, and 7.5 days, respectively. The final size matches that of AlphaZero and ELF.

The neural net has several output heads. Sampling positions from the self-play games, a \emph{policy} head predicts probable good moves while a \emph{game outcome value head} predicts if the game was ultimately won or lost. The loss function is:
\[ L = - c_{\text{g}} \sum_{r} z(r) \log(\hat{z}(r))  - \sum_{m} \pi(m) \log(\hat{\pi}(m)) + c_{L2} ||\theta||^2 \]
where $r \in \{\text{win},\text{loss}\}$ is the outcome for the current player, $z$ is a one-hot encoding of it, $\hat{z}$ is the neural net's prediction of $z$, $m$ ranges over the set of possible moves, $\pi$ is a target policy distribution derived from the playouts of the MCTS search, $\hat{\pi}$ is the prediction of $\pi$, $c_{L2}=\text{3e-5}$ sets an L2 penalty on the model parameters $\theta$, and $c_{\text{g}} = 1.5$ is a scaling constant. As described in later sections, we also add additional terms corresponding to other heads that predict auxiliary targets.

Training uses stochastic gradient descent with a momentum decay of 0.9 and a batch size of 256 (the largest size fitting on one GPU). It uses a fixed per-sample learning rate of 6e-5, except that the first 5 million samples (merely a few percent of the total steps) use a rate of 2e-5 to reduce instability from early large gradients. In KataGo's main run, the per-sample learning rate was also dropped to 6e-6 starting at about 17.5 days to maximize final strength. Samples are drawn uniformly from a growing moving window of the most recent data, with window size beginning at 250,000 samples and increasing to about 22 million by the end of the main run. See Appendix C for details. 

Training uses a version of \emph{stochastic weight averaging} \cite{SWA}. Every roughly 250,000 training samples, a snapshot of the weights is saved, and every four snapshots, a new candidate neural net is produced by taking an exponential moving average of snapshots with decay = 0.75 (averaging four snapshots of lookback). Candidate nets must pass a \emph{gating} test by winning at least 100 out of 200 test games against the current net to become the new net for self-play. See Appendix E for details. 

In total, KataGo's main run lasted for 19 days using a maximum of 28 V100 GPUs at any time (averaging 26-27) and generated about 241 million training samples across 4.2 million games. Self-play games used Tromp-Taylor rules \cite{TrompTaylor} modified to not require capturing stones within pass-alive-territory\footnotemark. ``Ko'', ``suicide'', and ``komi'' rules also varied from Tromp-Taylor randomly, and some proportion of games were randomly played on smaller boards\footnotemark. See Appendix D for other details. 

\addtocounter{footnote}{-2}
\stepcounter{footnote}\footnotetext{In Go, a version of Benson's algorithm \cite{Benson} can prove areas safe even given unboundedly many consecutive opponent moves (``pass-alive''), enabling this minor optimization.}
\stepcounter{footnote}\footnotetext{Almost all major AlphaZero reproductions in Go have been hardcoded to fixed board sizes and rules. Although not the focus of this paper, KataGo's randomization allows training a \emph{single} model that generalizes across all these variations.}

\section{Major General Improvements} \label{GeneralSection}

\subsection{Playout Cap Randomization} \label{PlayoutCapRandomization}
One of the major improvements in KataGo's training process over AlphaZero is to randomly vary the number of playouts on different turns to relieve a major tension between policy and value training.

In the AlphaZero process, the game outcome value target is highly data-limited, with only one noisy binary result per entire game. Holding compute fixed, it would likely be beneficial for value training to use only a small number of playouts per turn to generate more games, even if those games are of slightly lower quality. For example, in the first version of AlphaGo, self-play using only a single playout per turn (i.e., directly using the policy) was still of sufficient quality to train a decent value net \cite{AG}.

However, informal prior research by Forsten (2019) \cite{OptimalVisits} has suggested that at least in Go, ideal numbers of playouts for policy learning are much larger, not far from AlphaZero's choice of 800 playouts per move \cite{AZ}. Although the policy gets many samples per game, unless the number of playouts is larger than ideal for value training, the search usually does not deviate much from the policy prior, so the policy does not readily improve.

We introduce \emph{playout cap randomization} to mitigate this tension. On a small proportion $p$ of turns, we perform a full search, stopping when the tree reaches a cap of $N$ nodes, and for all other turns we perform a fast search with a much smaller cap of $n < N$. Only turns with a full search are recorded for training. For fast searches, we also disable Dirichlet noise and other explorative settings, maximizing strength. For KataGo's main 19-day run, we chose $p = 0.25$ and $(N,n) = (600,100)$ initially, annealing up to $(1000,200)$ after the first two days of training. 

Because most moves use a fast search, more games are played, improving value training. But since $n$ is small, fast searches cost only a limited fraction of the computation time, so the drop in the number of good policy samples per computation time is not large. The ablation studies presented in section \ref{AblationStudies} indicate that playout cap randomization indeed outperforms a variety of fixed numbers of playouts.

\subsection{Forced Playouts and Policy Target Pruning} \label{ForcedTPSection}
Like AlphaZero and other implementations such as ELF and Leela Zero, KataGo uses the final root playout distribution from MCTS to produce the policy target for training. However, KataGo does \emph{not} use the raw distribution. Instead, we introduce \emph{policy target pruning}, a new method which enables improved exploration via \emph{forced playouts}.

We observed in informal tests that even if a Dirichlet noise move was good, its initial evaluation might be negative, preventing further search and leaving the move undiscovered. Therefore, for each child $c$ of the root that has received any playouts, we ensure it receives a minimum number of \emph{forced playouts} based on the noised policy and the total sum of playouts so far:
\[ n_{\text{forced}}(c) = \left( k P(c) \sum_{c'} N(c') \right)^{1/2} \]
We do this by setting the MCTS selection urgency $\text{PUCT}(c)$ to infinity whenever a child of the root has fewer than this many playouts. The exponent of $1/2 < 1$ ensures that forced playouts scale with search but asymptotically decay to a zero proportion for bad moves, and $k = 2$ is large enough to actually force a small percent of playouts in practice.

\begin{figure}[!ht]
\centerline{
\includegraphics[width=2.5in]{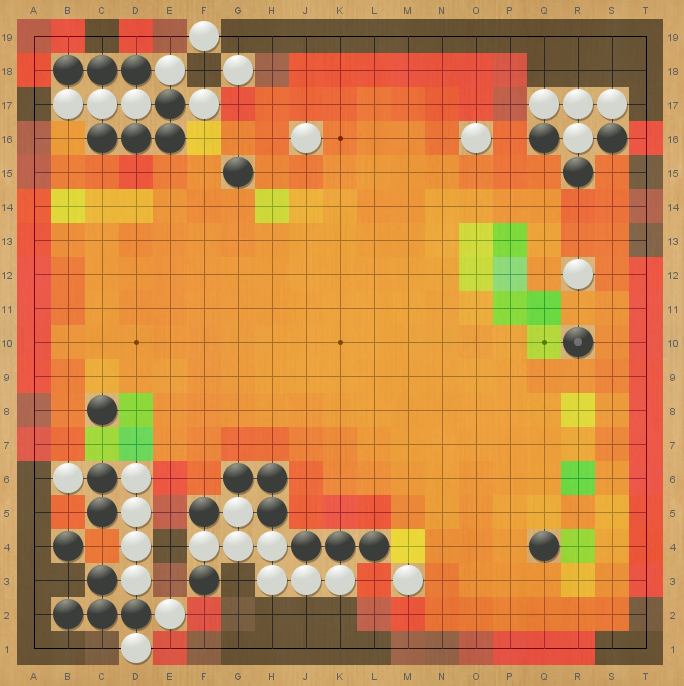}
\includegraphics[width=2.5in]{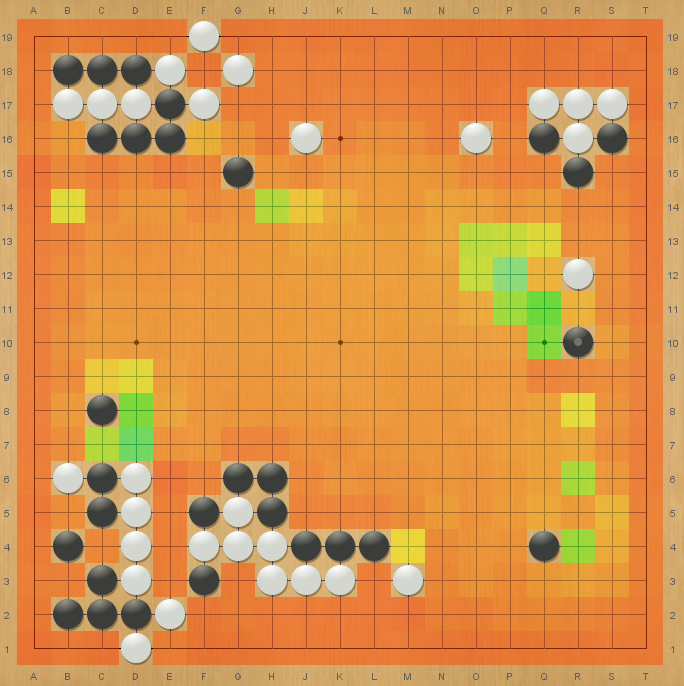}
}
\caption{Log policy of 10-block nets, white to play. Left: trained with forced playouts and policy target pruning. Right: trained without. Dark/red through bright green ranges from about $p=\text{2e-4}$ to $p=1$. Pruning reduces the policy mass on many bad moves near the edges.}
\label{Policy}
\end{figure}

However, the vast majority of the time, noise moves are bad moves, and in AlphaZero since the policy target is the playout distribution, we would train the policy to predict these extra bad playouts. Therefore, we perform a \emph{policy target pruning} step. In particular, we identify the child $c^*$ with the most playouts, and then from each other child $c$, we subtract up to $n_{\text{forced}}$ playouts so long as it does not cause $\text{PUCT}(c) >= \text{PUCT}(c^*)$, holding constant the \emph{final} utility estimate for both. This subtracts all ``extra'' playouts that normal PUCT would not have chosen on its own, unless a move was found to be good. Additionally, we outright prune children that are reduced to a single playout. See Figure \ref{Policy} for a visualization of the effect on the learned policy.

The critical feature of such pruning is that it allows \emph{decoupling the policy target in AlphaZero from the dynamics of MCTS or the use of explorative noise}. There is no reason to expect the optimal level of playout dispersion in MCTS to also be optimal for the policy target and the long-term convergence of the neural net. Our use of policy target pruning with forced playouts, though an improvement, is only a simple application of this method. We are eager to explore others in the future, including alterations to the PUCT formula itself\footnotemark.

\footnotetext{The PUCT formula $V(c) + c_{\text{PUCT}} P(c) \frac{ \sqrt{\sum_{c'} N(c')} } { 1 + N(c) }$ has the property that if $V$ is constant, then playouts will be roughly proportional to $P$. Informal tests suggest this is important to the convergence of $P$, and without something like target pruning, alternate formulas can disrupt training even when improving match strength.}


\subsection{Global Pooling} \label{GlobalPooling}
Another improvement in KataGo over earlier work is from adding \emph{global pooling} layers at various points in the neural network. This enables the convolutional layers to condition on global context, which would be hard or impossible with the limited perceptual radius of convolution alone.

In KataGo, given a set of $c$ channels, a \emph{global pooling layer} computes (1) the mean of each channel, (2) the mean of each channel scaled linearly with the width of the board, and (3) the maximum of each channel. This produces a total of $3c$ output values. These layers are used in a \emph{global pooling bias structure} consisting of:
\begin{itemize} 
\item Input tensors $X$ (shape $b \times b \times c_X$) and $G$ (shape $b \times b \times c_G$).
\item A batch normalization layer and ReLu activation applied to $G$ (output shape $b \times b \times c_G$).
\item A global pooling layer (output shape $3c_G$).
\item A fully connected layer to $c_X$ outputs (output shape $c_X$).
\item Channelwise addition with $X$ (output shape $b \times b \times c_X$).
\end{itemize}

\begin{figure}[ht]
\centerline{
\includegraphics[width=4.5in]{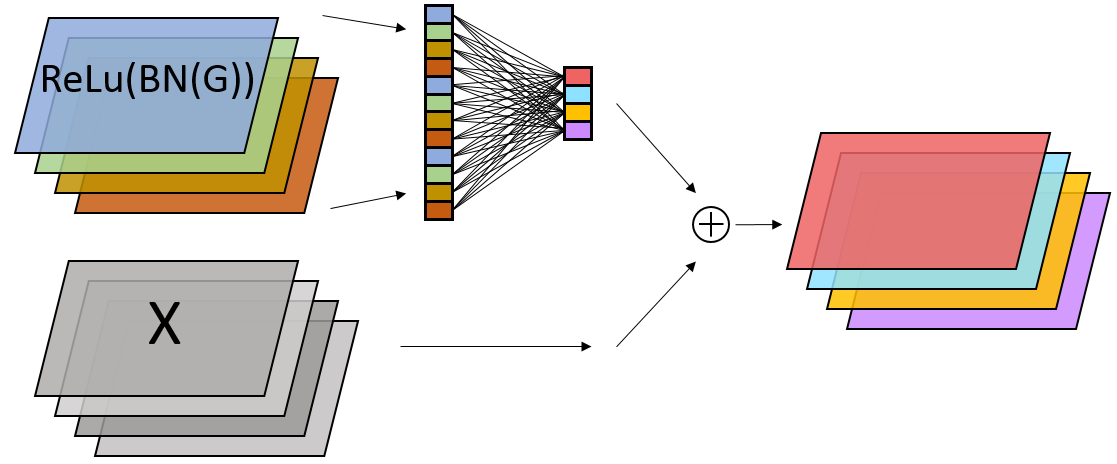}
}
\caption{Global pooling bias structure, globally aggregating values of one set of channels to bias another set of channels.}
\label{GPool}
\end{figure}

See Figure \ref{GPool} for a diagram. This structure follows the first convolution layer of two to three of the residual blocks in KataGo's neural nets, and the first convolution layer in the policy head. It is also used in the value head with a slight further modification. See Appendix A for details.

In Section \ref{AblationStudies} our experiments show that this greatly improves the later stages of training. As Go contains explicit nonlocal tactics (``ko''), this is not surprising. But global context should help even in domains without explicit nonlocal interactions. For example, in a wide variety of strategy games, strong players, when winning, alter their local move preferences to favor ``simple'' options, whereas when losing they seek ``complication''. Global pooling allows convolutional nets to internally condition on such global context.


The general idea of using global context is by no means novel to our work. For example, Hu et al. (2018) have introduced a ``Squeeze-and-Excitation'' architecture to achieve new results in image classification \cite{SE}. Although their implementation is different, the fundamental concept is the same. And though not formally published, Squeeze-Excite-like architectures are now in use in some online AlphaZero-related projects \cite{LeelaChessZero,MiniGo}, and we look forward to exploring it ourselves in future research.

\subsection{Auxiliary Policy Targets} \label{AuxiliaryPolicyTargets}
As another generalizable improvement over AlphaZero, we add an auxiliary policy target that predicts the opponent's reply on the following turn to improve regularization. This idea is not entirely new, having been found by Tian and Zhu in Facebook's bot Darkforest to improve supervised move prediction \cite{Darkforest}, but as far as we know, KataGo is the first to apply it to the AlphaZero process.

We simply have the policy head output a new channel predicting this target, adding a term to the loss function:
\[ - w_{\text{opp}} \sum_{m \in \text{moves}} \pi_{\text{opp}}(m) \log(\hat{\pi}_{\text{opp}}(m)) \]
where $\pi_{\text{opp}}$ is the policy target that will be recorded for the turn \emph{after} the current turn, $\hat{\pi}_{\text{opp}}$ is the neural net's prediction of $\pi_{\text{opp}}$, and $w_{\text{opp}} = 0.15$ weights this target only a fraction as much as the actual policy, since it is for regularization only and is never actually used for play.

We find in Section \ref{AblationStudies} that this produces a modest but clear benefit. Moreover, this idea could apply to a wide range of reinforcement-learning tasks. Even in single-agent situations, one could predict one's own future actions, or predict the environment (treating the environment as an ``agent''). Along with Section \ref{OwnershipAndScoreTargets}, it shows how enriching the training data with additional targets is valuable when data is limited or expensive. We believe it deserves attention as a simple and nearly costless method to regularize the AlphaZero process or other broader learning algorithms.

\section{Major Domain-Specific Improvements} \label{SpecificSection}

\subsection{Auxiliary Ownership and Score Targets} \label{OwnershipAndScoreTargets}
One of the major improvements in KataGo's training process over AlphaZero comes from adding auxiliary ownership and score prediction targets. Similar targets were earlier explored in work by Wu et al. (2018) \cite{MLVGo} in supervised learning, where the authors found improved mean squared error on human game result prediction and mildly improved the strength of their overall bot, CGI. 

To our knowledge, KataGo is the first to publicly apply such ideas to the reinforcement-learning-like context of self-play training in Go\footnotemark. While the targets themselves are game-specific, they also highlight a more general heuristic underemphasized in transfer- and multi-task-learning literature.

\footnotetext{A bot ``Golaxy'' developed by a Chinese research group appears also capable of making score predictions, but we are not currently aware of anywhere they have published their methods.}

As observed earlier, in AlphaZero, learning is highly constrained by data and noise on the game outcome prediction. But although the game outcome is noisy and binary, it is a direct function of finer variables: the final score difference and the ownership of each board location\footnotemark. Decomposing the game result into these finer variables and predicting them as well should improve regularization. 

Therefore, we add these outputs and three additional terms to the loss function:


\footnotetext{In Go, every point occupied or surrounded at the end of the game scores 1 point. The second player also receives a \emph{komi} of typically 7.5 points. The player with more points wins.}

\begin{itemize}
\item Ownership loss:
\[ - w_o \sum_{l \in \text{board}} \sum_{p \in \text{players}} o(l,p) \log \left(\hat{o}(l,p)\right) \]
where $o(l,p) \in \{0,0.5,1\}$ indicates if $l$ is finally owned by $p$, or is shared, $\hat{o}$ is the prediction of $o$, and $w_o = 1.5/b^2$ where $b \in [9,19]$ is the board width.

\item Score belief loss (``pdf''): 
\[ - w_{\text{spdf}} \sum_{x \in \text{possible scores}} p_s(x) \log(\hat{p}_s(x)) \]
where $p_s$ is a one-hot encoding of the final score difference, $\hat{p}_s$ is the prediction of $p_s$, and $w_{\text{spdf}} = 0.02$.

\item Score belief loss (``cdf''): 
\[ w_{\text{scdf}} \sum_{x \in \text{possible scores}} \left( \sum_{y < x} p_s(y) - \hat{p}_s(y) \right)^2 \]
where $w_{\text{scdf}} = 0.02$. While the ``pdf'' loss rewards guessing the score exactly, this ``cdf'' loss pushes the overall mass to be near the final score.
\end{itemize}

We show in our ablation runs in Section \ref{AblationStudies} that these auxiliary targets noticeably improve the efficiency of learning. This holds even up through the ends of those runs (though shorter, the runs still reach a strength similar to human-professional), well beyond where the neural net must have already developed a sophisticated judgment of the board. 

It might be surprising that these targets would continue to help beyond the earliest stages. We offer an intuition: consider the task of updating from a game primarily lost due to misjudging a particular region of the board. With only a final binary result, the neural net can only ``guess'' at what aspect of the board position caused the loss. By contrast, with an ownership target, the neural net receives direct feedback on which area of the board was mispredicted, with large errors and gradients localized to the mispredicted area. The neural net should therefore require fewer samples to perform the correct credit assignment and update correctly.

As with auxiliary policy targets, these results are consistent with work in transfer and multi-task learning showing that adding targets or tasks can improve performance. But the literature is scarcer in theory on \emph{when} additional targets may help -- see Zhang and Yang (2017) \cite{MultiSurvey} for discussion as well as Bingel and Søgaard (2017) \cite{MultiNLP} for a study in NLP domains. Our results suggest a heuristic: \emph{whenever a desired target can be expressed as a sum, conjunction, or disjunction of separate subevents, or would be highly correlated with such subevents, predicting those subevents is likely to help}. This is because such a relation should allow for a specific mechanism: that gradients from a mispredicted sub-event will provide sharper, more localized feedback than from the overall event, improving credit assignment.

\begin{figure}
\centerline{
\includegraphics[width=4.3in]{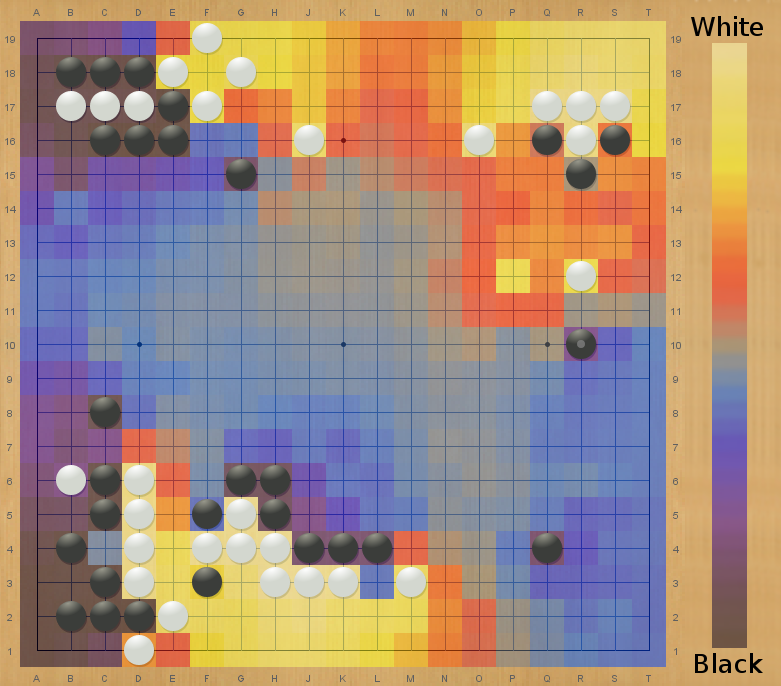}
}
\caption{Visualization of ownership predictions by the trained neural net.}
\label{OwnershipVis}
\end{figure}

We are likely not the first to discover such a heuristic. And of course, it may not always be applicable. But we feel it is worth highlighting both for practical use and as an avenue for further research, because when applicable, it is a potential path to study and improve the reliability of multi-task-learning approaches for more general problems.

\subsection{Game-specific Features}  \label{InputFeatures}
In addition to raw features indicating the stones on the board, the history, and the rules and komi in effect, KataGo includes a few game-specific higher-level features in the input to its neural net, similar to those in earlier work \cite{ClarkStorkey,Cazenave,DeepGoConv}. These features are liberties, komi parity, pass-alive regions, and features indicating ladders (a particular kind of capture tactic). See Appendix A for details.

Additionally, KataGo uses two minor Go-specific optimizations, where after a certain number of consecutive passes, moves in pass-alive territory are prohibited, and where a tiny bias is added to favor passing when passing and continuing play would lead to identical scores. Both optimizations slightly speed up the end of the game.

To measure the effect of these game-specific features and optimizations, we include in Section \ref{AblationStudies} an ablation run that disables both ending optimizations and all input features other than the locations of stones, previous move history, and game rules. We find they contribute noticeably to the learning speed, but account for only a small fraction of the total improvement in KataGo.


\section{Results} \label{Experiments}

\subsection{Testing Versus ELF and Leela Zero} \label{TestingVersusLeelaZero}

We tested KataGo against ELF and Leela Zero 0.17 using their publicly-available source code and trained networks.

We sampled roughly every fifth Leela Zero neural net over its training history from ``LZ30'' through ``LZ225'', the last several networks well exceeding even ELF's strength. Between every pair of Leela Zero nets fewer than 35 versions apart, we played about 45 games to establish approximate relative strengths of the Leela Zero nets as a benchmark.

We also sampled KataGo over its training history, for each version playing batches of games versus random Leela Zero nets with frequency proportional to the predicted variance $p(1-p)$ of the game result. The winning chance $p$ was continuously estimated from the global Bayesian maximum-likelihood Elo based on all game results so far\footnotemark. This ensured that games would be varied yet informative. We also ran ELF's final ``V2'' neural network using Leela Zero's engine\footnotemark, with ELF playing against both Leela Zero and KataGo using the same opponent sampling.

\addtocounter{footnote}{-2}
\stepcounter{footnote}\footnotetext{Using a custom implementation of BayesElo \cite{BayesElo}.}
\stepcounter{footnote}\footnotetext{ELF and Leela Zero neural nets are inter-compatible.}

Games used a 19x19 board with a fixed 7.5 komi under Tromp-Taylor rules, with a fixed 1600 visits, resignation below 2\% winrate, and multithreading disabled. To encourage opening variety, both bots randomized with a temperature of 0.2 in the first 20 turns. Both also used a ``lower-confidence-bound'' move selection method to improve match strength \cite{LCB}. Final Elo ratings were based on the final set of about 21000 games.


\begin{figure}
\centerline{
\includegraphics[width=4.8in]{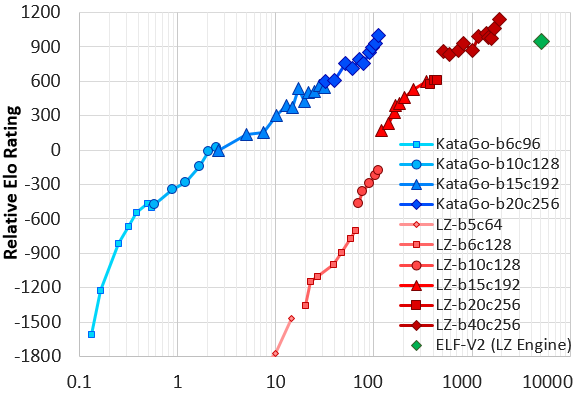}
}
\caption{1600-visit Elo progression of KataGo (blue, leftmost) vs. Leela Zero (red, center) and ELF (green diamond). X-axis: self-play cost in billions of equivalent 20 block x 256 channel queries. Note the log-scale. Leela Zero's costs are highly approximate.}
\label{PlotVsLZ}
\end{figure}

To compare the efficiency of training, we computed a crude indicative metric of total self-play computation by modeling a neural net with $b$ blocks and $c$ channels as having cost $\sim bc^2$ per query\footnotemark. For KataGo we just counted self-play queries for each size and multiplied. For ELF, we approximated queries by sampling the average game length from its public training data and multiplied by ELF's 1600 playouts per move, discounting by 20\% to roughly account for transposition caching. For Leela Zero we estimated it similarly, also interpolating costs where data was missing\footnotemark. Leela Zero also generated data using ELF's prototype networks, but we did \emph{not} attempt to estimate this cost\footnotemark.

\addtocounter{footnote}{-3}
\stepcounter{footnote}\footnotetext{This metric was chosen in part as a very rough way to normalize out hardware and engineering differences. For KataGo, we also conservatively computed costs under this metric as if all queries were on the full 19x19 board.}
\stepcounter{footnote}\footnotetext{Due to online hosting issues, some Leela Zero training data is no longer publicly available.}
\stepcounter{footnote}\footnotetext{At various points, Leela Zero also used data from stronger ELF OpenGo nets, likely causing it to learn faster than it would unaided. We did \emph{not} attempt to count the cost of this additional data.}

KataGo compares highly favorably with both ELF and Leela Zero. Shown in Figure \ref{PlotVsLZ} is a plot of Elo ratings versus estimated compute for all three. KataGo outperforms ELF in learning efficiency under this metric by about a factor of 50. Leela Zero appears to outperform ELF as well, but the Elo ratings would be expected to unfairly favor Leela since its final network size is 40 blocks, double that of ELF, and the ratings are based on equal search nodes rather than GPU cost. Additionally, Leela Zero's training occurred over multiple years rather than ELF's two weeks, reducing latency and parallelization overhead. Yet KataGo still outperforms Leela Zero by a factor of 10 despite the same network size as ELF and a similarly short training time. Early on, the improvement factor appears larger, but partly this is because the first 10\%-15\% of Leela Zero's run contained some bugs that slowed learning.

We also ran three 400-game matches on a single V100 GPU against ELF using ELF's native engine. In the first, both sides used 1600 playouts/move with no batching. In the second, KataGo used 9s/move (16 threads, max batch size 16) and ELF used 16,000 playouts/move (batch size 16), which ELF performs in 9 seconds. In the third, we doubled ELF's batch size, improving its nominal speed to 7.5s/move, and lowered KataGo to 7.5s/move. As summarized in Table \ref{VsELFTable}, in all three matches KataGo defeated ELF, confirming its strength level at both low and high playouts and at both fixed search and fixed wall clock time settings.

\begin{table}
\begin{center}
  \begin{tabular}{| l | c | c |}
    \hline
    Match Settings & Wins v ELF & Elo Diff \\ \hline
    1600 playouts/mv no batching & 239 / 400 & 69 $\pm$ 36  \\ 
    9.0 secs/mv, ELF batchsize 16 & 246 / 400 & 81 $\pm$ 36 \\ 
    7.5 secs/mv, ELF batchsize 32 & 254 / 400 & 96 $\pm$ 37 \\ \hline
  \end{tabular}
\end{center}
\caption{KataGo match results versus ELF, with the implied Elo difference (plus or minus two std. deviations of confidence).}
\label{VsELFTable}
\end{table}


\subsection{Ablation Runs} \label{AblationStudies}

To study the impact of the techniques presented in this paper, we ran shorter training runs with various components removed. These ablation runs went for about 2 days each, with identical parameters except for the following differences:
\begin{itemize}
\item FixedN - Replaces playout cap randomization with a fixed cap $N \in \{100,150,200,250,600\}$. For $N=600$ the window size was also doubled, as an informal test without doubling showed major overfitting due to lack of data.
\item NoForcedTP - Removes forced playouts and policy target pruning.
\item NoGPool - Removes global pooling from residual blocks and the policy head except for computing the ``pass'' output.
\item NoPAux - Removes the auxiliary policy target.
\item NoVAux - Removes the ownership and score targets.
\item NoGoFeat - Removes all game-specific higher-level input features and the minor optimizations involving passing listed in Section \ref{InputFeatures}.
\end{itemize}

We sampled neural nets from these runs together with KataGo's main run, and evaluated them the same way as when testing against Leela Zero and ELF: playing 19x19 games between random versions based on the predicted variance $p(1-p)$ of the result. Final Elos are based on the final set of about 147,000 games (note that these Elos are not directly comparable to those in Section \ref{TestingVersusLeelaZero}).

\begin{figure}[!ht]
\centerline{
\includegraphics[width=4.5in]{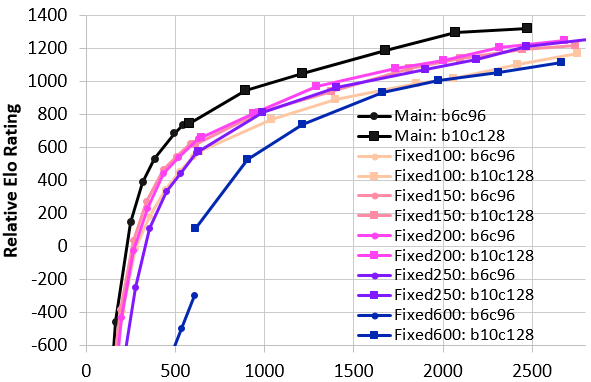}
}
\caption{KataGo's main run versus Fixed runs. X-axis is the cumulative self-play cost in millions of equivalent 20 block x 256 channel queries.}
\label{AblateOsc}
\end{figure}

\begin{figure}
\centerline{
\includegraphics[width=4.5in]{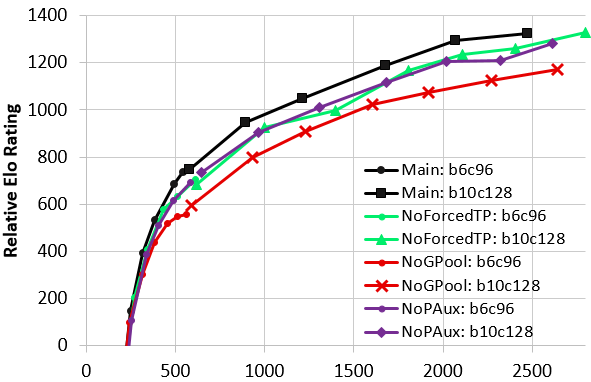}
}
\caption{KataGo's main run versus NoGPool, NoForcedTP, NoPAux. X-axis is the cumulative self-play cost in millions of equivalent 20 block x 256 channel queries.}
\label{AblateGPool}
\end{figure}

\begin{figure}
\centerline{
\includegraphics[width=4.5in]{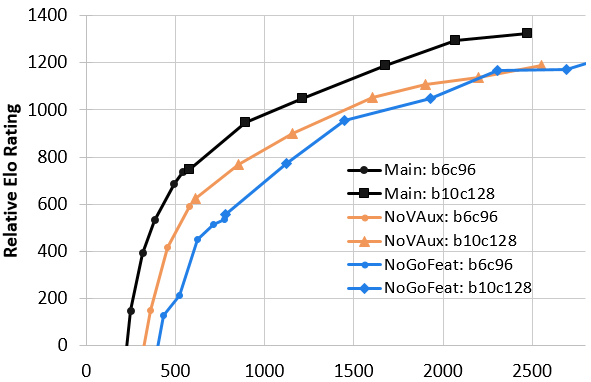}
}
\caption{KataGo's main run versus NoVAux, NoGoFeat. X-axis is the cumulative self-play cost in millions of equivalent 20 block x 256 channel queries.}
\label{AblateAux}
\end{figure}

As shown in Figure \ref{AblateOsc}, playout cap randomization clearly outperforms a wide variety of possible fixed values of playouts. This is precisely what one would expect if the technique relieves the tension between the value and policy targets present for any fixed number of playouts. Interestingly, the 600-playout run showed a large jump in strength when increasing neural net size. We suspect this is due to poor convergence from early overfitting not entirely mitigated by doubling the training window.

As shown in Figure \ref{AblateGPool}, global pooling noticeably improved learning efficiency, and forced playouts with policy target pruning and auxiliary policy targets also provided smaller but clear gains. Interestingly, all three showed little effect early on compared to later in the run. We suspect their relative value continues to increase beyond the two-day mark at which we stopped the ablation runs. These plots suggest that the total value of these general enhancements to self-play learning, along with playout cap randomization, is large.

As shown in Figure \ref{AblateAux}, removing auxiliary ownership and score targets resulted in a noticeable drop in learning efficiency. These results confirm the value of these auxiliary targets and the value, at least in Go, of regularization by predicting subcomponents of targets. Also, we observe a drop in efficiency from removing Go-specific input features and optimizations, demonstrating that there is still significant value in such domain-specific methods, but also accounting for only a part of the total speedup achieved by KataGo. 

See Table \ref{AblateTable} for a summary. The product of the acceleration factors shown is approximately 9.1x. We suspect this is an underestimate of the true speedup since several techniques continued to increase in effectiveness as their runs progressed and the ablation runs were shorter than our full run. Some remaining differences with ELF and/or AlphaZero are likely due to infrastructure and implementation. Unfortunately, it was beyond our resources to replicate ELF and/or AlphaZero's infrastructure of thousands of GPUs/TPUs for a precise comparison, or to run more extensive ablation runs each for as long as would be ideal.

\begin{table}[h]
\begin{center}
  \begin{tabular}{| l | c | c | c | c |}
    \hline
    Removed Component & Elo & Factor \\ \hline 
    (Main Run, baseline) & 1329 & 1.00x \\ 
    Playout Cap Randomization & 1242 & 1.37x  \\ 
    F.P. and Policy Target Pruning & 1276 & 1.25x \\ 
    Global Pooling & 1153 & 1.60x \\ 
    Auxiliary Policy Targets & 1255 & 1.30x \\ 
    Aux Owner and Score Targets & 1139 & 1.65x \\ 
    Game-specific Features and Opts & 1168 & 1.55x \\ \hline
  \end{tabular}
\end{center}
\caption{For each technique, the Elo of the ablation run omitting it as of reaching 2.5G equivalent 20b x 256c self-play queries ($\approx$ 2 days), and the factor increase in training time to reach that Elo. Factors are \emph{approximate} and are based on shorter runs.}
\label{AblateTable}
\end{table}

\section{Conclusions And Future Work} \label{Conclusions And Future Work}
Still beginning only from random play with no external data, our bot KataGo achieves a level competitive with some of the top AlphaZero replications, but with an enormously greater efficiency than all such earlier work. In this paper, we presented a variety of techniques we used to improve self-play learning, many of which could be readily applied to other games or to problems in reinforcement learning more generally. Furthermore, our domain-specific improvements demonstrate a remaining gap between basic AlphaZero-like training and what could be possible, while also suggesting principles and possible avenues for improvement in general methods. We hope our work lays a foundation for further improvements in the data efficiency of reinforcement learning.

\bibliographystyle{plain.bst}
\bibliography{Accelerating_Self_Play_Learning_In_Go_2020}

\begin{appendices}

\pagebreak

\section{Neural Net Inputs and Architecture} \label{NNArchitecture}
The following is a detailed breakdown of KataGo's neural net inputs and architecture. The neural net has two input tensors, which feed into a \emph{trunk} of residual blocks. Attached to the end of the trunk are a \emph{policy head} and a \emph{value head}, each with several outputs and subcomponents.

\subsection{Inputs}
The input to the neural net consists of two tensors, a $b \times b \times 18$ tensor of $18$ binary features for each board location where where $b \in [b_{\text{min}}, b_{\text{max}}] = [9,19]$ is the width of the board, and a vector with $10$ real values indicating overall properties of the game state. These features are summarized in Tables \ref{InputBinaryFeaturesTable} and \ref{InputGlobalFeaturesTable}

\begin{table}[h]
\begin{center}
  \begin{tabular}{| c | l |}
    \hline
    \# Channels & Feature  \\ \hline
    1 & Location is on board \\ \hline
    2 & Location has \{own,opponent\} stone \\ \hline
    3 & Location has stone with \{1,2,3\} liberties \\ \hline
    1 & Moving here illegal due to ko/superko \\ \hline
    5 & The last 5 move locations, one-hot \\ \hline
    3 & Ladderable stones \{0,1,2\} turns ago \\ \hline
    1 & Moving here catches opponent in ladder \\ \hline
    2 & Pass-alive area for \{self,opponent\} \\ \hline
  \end{tabular}
\end{center}
\caption{Binary spatial-varying input features to the neural net. A ``ladder'' occurs when stones are forcibly capturable via consecutive inescapable atari (i.e. repeated capture threat).}
\label{InputBinaryFeaturesTable}
\end{table}

\begin{table}[h]
\begin{center}
  \begin{tabular}{| c | l |}
    \hline
    \# Channels & Feature \\ \hline
    5 & Which of the previous 5 moves were pass? \\ \hline
    1 & Komi / 15.0 (current player's perspective) \\ \hline
    2 & Ko rules (simple,positional,situational) \\ \hline
    1 & Suicide allowed? \\ \hline
    1 & Komi + board size parity \\ \hline
  \end{tabular}
\end{center}
\caption{Overall game state input features to the neural net.}
\label{InputGlobalFeaturesTable}
\end{table}

\subsection{Global Pooling}
Certain layers in the neural net are \emph{global pooling layers}. Given a set of $c$ channels, a \emph{global pooling layer} computes:
\begin{enumerate}
\item The mean of each channel 
\item The mean of each channel multiplied by $\frac{1}{10}(b - b_{\text{avg}})$
\item The maximum of each channel. 
\end{enumerate}
where $b_{\text{avg}} = 0.5 (b_{\text{min}} + b_{\text{max}}) = 0.5 (9 + 19)$. This produces a total of $3c$ output values. The multiplication in (2) allows training weights that work across multiple board sizes, and the subtraction of $b_{\text{avg}}$ and scaling by $1/10$ improve orthogonality and ensure values remain near unit scale. In the \emph{value head}, (3) is replaced with the mean of each channel multiplied by $\frac{1}{100}((b - b_{\text{avg}})^2 - \sigma^2)$ where $\sigma^2 = \frac{1}{11}\sum_{b'=9}^{19} (b'-b_{\text{avg}})^2$. This is since the value head computes values like score difference that need to scale quadratically with board width. As before, subtracting $\sigma^2$ and scaling improves orthogonality and normality.

Using such layers, a \emph{global pooling bias structure} takes input tensors $X$ (shape $b \times b \times c_X$) and $G$ (shape $b \times b \times c_G$) and consists of:
\begin{itemize}
\item A batch normalization layer and ReLU activation applied to $G$ (output shape $b \times b \times c_G$).
\item A global pooling layer (output shape $3c_G$).
\item A fully connected layer to $c_X$ outputs (output shape $c_X$).
\item Channelwise addition with $X$, treating the $c_X$ values as per-channel biases (output shape $b \times b \times c_X$).
\end{itemize}

\subsection{Trunk}

The \emph{trunk} consists of:
\begin{itemize}
\item A 5x5 convolution of the binary spatial input tensor outputting $c$ channels. In parallel, a fully connected linear layer on the overall game state input tensor outputting $c$ channels, producing biases that are added channelwise to the result of the 5x5 convolution.
\item A stack of $n$ residual blocks. All but two or three of the blocks are ordinary pre-activation ResNet blocks, consisting of the following in order:
  \begin{itemize}
  \item A batch-normalization layer.
  \item A ReLU activation function.
  \item A 3x3 convolution outputting $c$ channels.
  \item A batch-normalization layer.
  \item A ReLU activation function.
  \item A 3x3 convolution outputting $c$ channels.
  \item A skip connection adding the convolution result elementwise to the input to the block.
  \end{itemize}
\item The remaining two or three blocks, spaced at regular intervals in the trunk, use global pooling, consisting of the following in order:
  \begin{itemize}
  \item A batch-normalization layer.
  \item A ReLU activation function.
  \item A 3x3 convolution outputting $c$ channels.
  \item A \emph{global pooling bias structure} pooling the first $c_{\text{pool}}$ channels to bias the other $c-c_{\text{pool}}$ channels.
  \item A batch-normalization layer.
  \item A ReLU activation function.
  \item A 3x3 convolution outputting $c$ channels.
  \item A skip connection adding the convolution result elementwise to the input to the block.
  \end{itemize}
\item At the end of the trunk, a batch-normalization layer and one more ReLU activation function.
\end{itemize}

\subsection{Policy Head}

The \emph{policy head} consists of:
\begin{itemize}
\item A 1x1 convolution outputting $c_{\text{head}}$ channels (``$P$'') and in parallel a 1x1 convolution outputting $c_{\text{head}}$ channels (``$G$'').
\item A \emph{global pooling bias structure} pooling the output of $G$ to bias the output of $P$.
\item A batch-normalization layer.
\item A ReLU activation function.
\item A 1x1 convolution with $2$ channels, outputting two policy distributions in logits over moves on each of the locations of the board. The first channel is the predicted policy $\hat{\pi}$ for the current player. The second channel is the predicted policy $\hat{\pi}_{\text{opp}}$ for \emph{the opposing player on the subsequent turn}.
\item In parallel, a fully connected linear layer from the globally pooled values of $G$ outputting $2$ values, which are the logits for the two policy distributions for making the pass move for $\hat{\pi}$ and $\hat{\pi}_{\text{opp}}$, as the pass move is not associated with any board location.
\end{itemize}

\subsection{Value Head}

The \emph{value head} consists of:
\begin{itemize}
\item A 1x1 convolution outputting $c_{\text{head}}$ channels (``$V$'').
\item A \emph{global pooling layer} of $V$ outputting $3c_{\text{head}}$ values (``$V_{\text{pooled}}$'').

\item A game-outcome subhead consisting of:
\begin{itemize}
  \item A fully-connected layer from $V_{\text{pooled}}$ including bias terms outputting $c_{\text{val}}$ values.
  \item A ReLU activation function.
  \item A fully-connected layer from $V_{\text{pooled}}$ including bias terms outputting $9$ values. 
  \begin{itemize}
    \item The first $3$ values are a distribution in logits whose softmax $\hat{z}$ predicts among the three possible game outcomes \emph{win}, \emph{loss}, and \emph{no result} (the latter being possible under non-superko rulesets in case of long-cycles).
    \item The fourth value is multiplied by 20 to produce a prediction $\hat{\mu}_s$ of the final score difference of the game in points\footnotemark.
    \item The fifth value has a softplus activation applied and is then multiplied by 20 to produce an estimate $\hat{\sigma}_s$ of the standard deviation of the predicted final score difference in points.
    \item The sixth through ninth values have a softplus activation applied are predictions $\hat{\text{rv}}_i$ of the expected variance in the MCTS root value for different numbers of playouts\footnotemark.
    \item All predictions are from the perspective of the current player.
  \end{itemize}
\end{itemize}

\addtocounter{footnote}{-2}
\stepcounter{footnote}\footnotetext{20 was chosen as an arbitrary reasonable scaling factor so that on typical data the neural net would only need to output values around unit scale, rather than tens or hundreds.}
\stepcounter{footnote}\footnotetext{In training the weight on this head is negligibly small. It is included only to enable future research on whether MCTS can be improved by biasing search towards more ``uncertain'' subtrees.}

\item An ownership subhead consisting of:
\begin{itemize}
  \item A 1x1 convolution of $V$ outputting 1 channel.
  \item A tanh activation function. 
  \item The result is a prediction $\hat{o}$ of the expected ownership of each location on the board, where $1$ indicates ownership by the current player and $-1$ indicates ownership by the opponent.
\end{itemize}

\item A final-score-distribution subhead consisting of:
\begin{itemize}
  \item A scaling component:
  \begin{itemize}
    \item A fully-connected layer from $V_{\text{}pooled}$ including bias terms outputting $c_{\text{val}}$ values.
    \item A ReLU activation function.
    \item A fully-connected layer including bias terms outputting $1$ value (``$\gamma$'').
  \end{itemize}
  \item For each possible final score value $s$: 
\[ s \in \{-S+0.5,-S+1.5,\dots,S-1.5,S-0.5 \} \]
where $S$ is a an upper bound for the plausible final score difference of any game\footnotemark, in parallel:
  \begin{itemize}  
    \item The $3c_{\text{head}}$ values from $V_{\text{}pooled}$ are concatenated with two additional values:
          \[(0.05 * s, \text{Parity}(s)-0.5)\]
          $0.05$ is an arbitrary reasonable scaling factor so that these values vary closer to unit scale. $\text{Parity}(s)$ is the binary indicator of whether a score value is normally possible or not due to parity of the board size and komi\footnotemark.
    \item A fully-connected layer (sharing weights across all $s$) from the $3c_{\text{head}} + 2$ values including bias terms outputting $c_{\text{val}}$ values.
    \item A ReLU activation function.
    \item A fully-connected layer (sharing weights across all $s$) from $V_{\text{pooled}}$ including bias terms, outputting $1$ value. 
  \end{itemize}
  \item The resulting $2S$ values multiplied by $\text{softplus}(\gamma)$ are a distribution in logits whose softmax $\hat{p}_s$ predicts the final score difference of the game in points. All predictions are from the perspective of the current player.
\end{itemize}

\end{itemize}

\addtocounter{footnote}{-2}
\stepcounter{footnote}\footnotetext{We use $S = 19*19 + 60$, since $19$ is the largest standard board size, and the extra $60$ conservatively allows for the possibility that the winning player wins all of the board \emph{and} has a large number of points from \emph{komi}.}
\stepcounter{footnote}\footnotetext{In Go, usually every point on the board is owned by one player or the other in a finished game, so the final score difference varies only in increments of 2 and half of values only rarely occur. Such a parity component is very hard for a neural net to learn on its own. But this feature is mostly for cosmetic purposes, omitting it should have little effect on overall strength).}

\subsection{Neural Net Parameters}

Four different neural net sizes were used in our experiments. Table \ref{NNConstants} summarizes the constants for each size. Additionally, the four different sizes used, respectively, 2, 2, 2, and 3 global pooling residual blocks in place of ordinary residual blocks, at regularly spaced intervals.

\begin{table}[!h]
\begin{center}
    \begin{tabular}{| l | c | c | c | c |}
        \hline
    Size   & b6$\times$c96 & b10$\times$c128 & b15$\times$c192 & b20$\times$c256 \\ \hline
    $n$    & 6  & 10 & 15 & 20 \\ \hline
    $c$    & 96 & 128 & 192 & 256 \\ \hline
    $c_{\text{pool}}$  & 32  & 32 & 64 & 64 \\ \hline
    $c_{\text{head}}$ & 32  & 32 & 32 & 48 \\ \hline
    $c_{\text{val}}$ & 48  & 64 & 80 & 96 \\ \hline
    
  \end{tabular}
\end{center}
\caption{Architectural constants for various neural net sizes.}
\label{NNConstants}
\end{table}

\pagebreak

\section{Loss Function} \label{LossFunction}

The loss function used for neural net training in KataGo is the sum of:
\begin{itemize}
\item Game outcome value loss:
\[ c_{\text{value}} \sum_{r \in \{\text{win},\text{loss}\}} z(r) \log(\hat{z}(r)) \]
where $z$ is a one-hot encoding of whether the game was won or lost by the current player, $\hat{z}$ is the neural net's prediction of $z$, and $c_{\text{value}} = 1.5$.

\item Policy loss:
\[ - \sum_{m \in \text{moves}} \pi(m) \log(\hat{\pi}(m)) \]
where $\pi$ is the target policy distribution and $\hat{\pi}$ is the prediction of $\pi$.

\item Opponent policy loss:
\[ - w_{\text{opp}} \sum_{m \in \text{moves}} \pi_{\text{opp}}(m) \log(\hat{\pi}_{\text{opp}}(m)) \]
where $\pi_{\text{opp}}$ is the target opponent policy distribution, $\hat{\pi}_{\text{opp}}$ is the prediction of $\pi_{\text{opp}}$, and $w_{\text{opp}} = 0.15$.

\item Ownership loss:
\[ - w_o \sum_{l \in \text{board}} \sum_{p \in \text{players}} o(l,p) \log \left(\hat{o}(l,p)\right) \]
where $o(l,p) \in \{0,0.5,1\}$ indicates if $l$ is finally owned by $p$, or is shared, $\hat{o}$ is the prediction of $o$, and $w_o = 1.5/b^2$ where $b \in [9,19]$ is the board width.

\item Score belief loss (``pdf''): 
\[ - w_{\text{spdf}} \sum_{x \in \text{possible scores}} p_s(x) \log(\hat{p}_s(x)) \]
where $p_s$ is a one-hot encoding of the final score difference, $\hat{p}_s$ is the prediction of $p_s$, and $w_{\text{spdf}} = 0.02$.

\item Score belief loss (``cdf''): 
\[ w_{\text{scdf}} \sum_{x \in \text{possible scores}} \left( \sum_{y < x} p_s(y) - \hat{p}_s(y) \right)^2 \]
where $w_{\text{scdf}} = 0.02$.

\item Score belief mean self-prediction:
\[ - w_{\text{sbreg}} \text{Huber}(\hat{\mu}_s - \mu_s,\,\delta = 10.0) \]
where $w_{\text{sbreg}} = 0.004$ and
\[ \mu_s = \sum_{x} x \hat{p}_s(x) \]
and $\text{Huber}(x,\delta)$ is the \emph{Huber loss function} equal to the squared error loss $f(x) = 1/2 x^2$ except that for $|x| > \delta$, instead $\text{Huber}(x,\delta) = f(\delta) + (|x|-\delta)\frac{df}{dx}(\delta)$. This avoids some cases of divergence in training due to large errors just after initialization, but otherwise is exactly identical to a plain squared error beyond the earliest steps of training.

Note that neural net is predicting itself - i.e. this is a regularization term for an otherwise unanchored output $\hat{\mu}_s$ to roughly equal to the mean score implied by the neural net's full score belief distribution. The neural net easily learns to make this output highly consistent with its own score belief\footnotemark. \addtocounter{footnote}{-1}

\item Score belief standard deviation self-prediction:
\[ - w_{\text{sbreg}} \text{Huber}(\hat{\sigma}_s - \sigma_s,\,\delta = 10.0) \]
where 
\[ \sigma_s = \left(\,\sum_{x} (x-\mu)^2 \hat{p}_s(x) \,\right)^{1/2} \]
Similarly, the neural net is predicting itself - i.e. this is a regularization term for an otherwise unanchored output $\hat{\sigma}_s$ to roughly equal to the standard deviation of the neural net's full score belief distribution. The neural net easily learns to make this output highly consistent with its own score belief\footnotemark.

\item Score belief scaling penalty:
\[ w_{\text{scale}} \gamma^2 \]
where $\gamma$ is the activation strength of the internal scaling of the score belief and $w_{\text{scale}} = 0.0005$. This prevents some cases of training instability involving the multiplicative behavior of $\gamma$ on the belief confidence where $\gamma$ grows too large, but otherwise should have little overall effect on training.

\item L2 penalty:
\[ c ||\theta||^2 \]
where $\theta$ are the model parameters and $c = 0.00003$, so as to bound the weight scale and ensure that the effective learning rate does not decay due to batch normalization's inability to constrain weight magnitudes.

\end{itemize}

KataGo also implements a term for predicting the variance of the MCTS root value intended for future MCTS research, but in all cases this term was used only with negligible or zero weight.

The coefficients on these new auxiliary loss terms were mostly guesses chosen so that empirical observed average gradients and loss values from them in training would be, e.g. anywhere from ten to forty percent as large as those from the main policy and value head terms - neither too small to affect training, nor too large and exceeding them. Beyond these initial guessed weights, they were NOT carefully tuned, since we could afford only a limited number of test runs. Although better tuning would likely help, such arbitrary reasonable values already appeared to give immediate and significant improvements.

\footnotetext{KataGo's play engine uses a separate GPU implementation so as to run independently of TensorFlow, and these self-prediction outputs allow convenient access to the mean and variance without needing to re-implement the score belief head. Also for technical reasons relating to tree re-use, using only the first and second moments instead of the full distribution is convenient.}

\pagebreak

\section{Training Details} \label{TrainingDetails}

In total, KataGo's main run lasted for 19 days using 16 V100 GPUs for self-play for the first two days and increasing to 24 V100 GPUs afterwards, and 2 V100 GPUs for gating, one V100 GPU for neural net training, and additionally one V100 GPU for neural net training when running the next larger size concurrently on the same data. It generated about 241 million training samples across 4.2 million games, across four neural net sizes, as summarized in Tables \ref{TrainSummaryTable} and \ref{TrainSummaryTable2}. 

\begin{table}[!h]
\begin{center}
  \begin{tabular}{| l | c | c | c | c |}
    \hline
    Size & Days & Train Steps & Samples & Games \\ \hline
    b6$\times$c96 & 0.75 & 98M & 23M & 0.4M \\ \hline
    b10$\times$c128 & 1.75 & 209M & 55M & 1.0M \\ \hline
    b15$\times$c192 & 7.5 & 506M & 140M &  2.5M \\ \hline
    b20$\times$c256 & 19 & 954M & 241M & 4.2M \\ \hline
  \end{tabular}
\end{center}
\caption{Training time of the strongest neural net of each size in KataGo's main run. ``Days'' is the time of finishing a size and switching to the next larger size , ``Train Steps'' indicates cumulative gradient steps taken measured in samples, ``Samples'' and ``Games'' indicate cumulative self-play data samples and games generated. }
\label{TrainSummaryTable}
\end{table}

\begin{table}[!h]
\begin{center}
  \begin{tabular}{| l | c | c |}
    \hline
    Size & Elo vs LZ/ELF & Rough strength \\ \hline
    b6$\times$c96  & -1276 & Strong/Top Amateur \\ \hline
    b10$\times$c128  & -850 & Strong Professional \\ \hline
    b15$\times$c192 & -329 & Superhuman \\ \hline
    b20$\times$c256  & +76 & Superhuman \\ \hline
  \end{tabular}
\end{center}
\caption{Approximate strength of the strongest neural net of each size in KataGo's main run at a search tree node cap of 1600. Elo values are versus a mix of various Leela Zero versions and ELF, anchored so that ELF is about Elo 0. }
\label{TrainSummaryTable2}
\end{table}

Training used a batch size of 256 and a per-sample learning rate of $6*10^{-5}$, or a per-batch learning rate of $256 * 6 * 10^{-5}$. However, the learning rate was lowered by a factor of $3$ for the first five million samples of training steps for each neural net to reduce early training instability, as well as lowered by a factor of $10$ for the final b20$\times$c256 net after 17.5 days of training for final tuning.

Training samples were drawn uniformly from a moving window of the most recent $N_{\text{window}}$ samples, where
\[ N_{\text{window}} = c \left( 1 + \beta \frac{ ( N_{\text{total}} / c ) ^ \alpha - 1} { \alpha } \right) \]
where $N_{\text{total}}$ is the total number of training samples generated in the run so far, $c = \text{250,000}$ and $\alpha = 0.75$ and $\beta = 0.4$. Though appearing complex, this is simply the sublinear curve $f(n) = n ^ \alpha$ but rescaled so that $f(c) = c$ and $f'(c) = \beta$.

\pagebreak

\section{Game Randomization and Termination} \label{GameInit}

KataGo randomizes in a variety of ways to ensure diverse training data so as to generalize across a wide range of rulesets, board sizes, and extreme match conditions, including handicap games and positions arising from mistakes or alternative moves in human games that would not occur in self-play. 

\begin{itemize}
\item Games are randomized uniformly between positional versus situational superko rules, and between suicide moves allowed versus disallowed.
\item Games are randomized in board size, with 37.5\% of games on 19x19 and increasing in KataGo's main run to 50\% of games after two days of training. The remaining games are triangularly distributed from 9x9 to 18x18, with frequency proportional to $1,2,\dots,10$.
\item Rather than using only a standard komi of $7.5$, komi is randomized by drawing from a normal distribution with mean $7$ and standard deviation $1$, truncated to 3 standard deviations and rounding to the nearest integer or half-integer. However, $5\%$ of the time, a standard deviation of $10$ is used instead, to give experience with highly unusual values of komi.
\item To enable experience with handicap game positions, $5\%$ of games are played as handicap games, where Black gets a random number of additional free moves at the start of the game, chosen randomly using the raw policy probabilities. Of those games, $90\%$ adjust komi to compensate White for Black's advantage based on the neural net's predicted final score difference. The maximum number of free Black moves is $0$ (no handicap) for board sizes 9 and 10, $1$ for board sizes 11 to 14, $2$ for board sizes 15 to 18, and $3$ for board size 19.
\item To initialize each game and ensure opening variety, the first $r$ moves of a game are played randomly directly proportionally to the raw policy distribution of the net, where $r$ is drawn from an exponential distribution with mean $0.04 * b^2$. where $b$ is the width of the board, and during the game, moves are selected proportionally to the target-pruned MCTS playout distribution raised to the power of $1/T$ where $T$ is a temperature constant. $T$ begins at $0.8$ and decays smoothly to $0.2$, with a halflife in turns equal to the width of the board $b$. These achieve essentially the same result to AlphaZero or ELF's temperature scaling in the first 30 moves of the game, except scaling with board size and varying more smoothly.
\item In $2.5\%$ of positions, the game is branched to try an alternative move drawn randomly from the policy of the net $70\%$ of the time with temperature $1$, $25\%$ of the time with temperature $2$, and otherwise with temperature infinity. A full search is performed to produce a policy training sample (the $MCTS$ search winrate is used for the game outcome target and the score and ownership targets are left unconstrained). This ensures that there is a small percentage of training data on how to respond to or refute moves that a full search might not play. Recursively, a random quarter of these branches are continued for an additional move.
\item In $5\%$ of games, the game is branched after the first $r$ turns where $r$ is drawn from an exponential distribution with mean $0.025 *b^2$. Between $3$ and $10$ moves are chosen uniformly at random, each given a single neural net evaluation, and the best one is played. Komi is adjusted to be fair. The game is then played to completion as normal. This ensures that there is always a small percentage of games with highly unusual openings.
\end{itemize}

Except for introducing a minimum necessary amount of entropy, the above settings very likely have only a limited effect on overall learning efficiency and strength. They were used primarily so that KataGo would have experience with alternate rules, komi values, handicap openings, and positions where both sides have played highly suboptimally in ways that would never normally occur in high-level play, making it more effective as a tool for human amateur game analysis.

Additionally, unlike in AlphaZero or in ELF, games are played to completion without resignation. However, during self-play if for 5 consecutive turns, the MCTS winrate estimate $p$ for the losing side has been less than 5\%, then to finish the game faster the number of visits is capped to $\lambda n + (1-\lambda)N$ where $n$ and $N$ are the small and large limits used in playout cap randomization and $\lambda = p / 0.05$ is the proportion of the way that $p$ is from 5\% to 0\%. Additionally, training samples are recorded with only $0.1 + 0.9 \,\lambda$ probability, stochastically downweighting training on positions where AlphaZero would have resigned. 

Relative to resignation, continuing play with reduced visit caps costs only slightly more but results in cleaner and less biased training targets, reduces infrastructural complexity such as monitoring for the rate of incorrect resignations, and enables the final ownership and final score targets to be easily computed. Since KataGo secondarily optimizes for score rather than just win/loss (see Appendix \ref{ScoreMaximization}), continued play itself also still provides some learning value since optimizing score can give a good signal even in won/lost positions.

\pagebreak

\section{Gating} \label{GatingAppendix}

Similar to AlphaGoZero, candidate neural nets must pass a \emph{gating} test to become the new net for self-play. Gating in KataGo is fairly lightweight - candidates need only win at least 100 out of 200 games against the current self-play neural net. Gating games use a fixed cap of 300 search tree nodes (increasing in KataGo's main run to 400 after 2 days), with the following parameter changes to minimize noise and maximize performance:

\begin{itemize}
\item The rules and board size are still randomized but komi is not randomized and is fixed at $7.5$.
\item Handicap games and branching are disabled. 
\item From the first turn, moves are played using full search rather than using the raw policy to play some of the first moves.
\item The temperature $T$ for selecting a move based on the MCTS playout distribution starts at $0.5$ instead of $0.8$.
\item Dirichlet noise and forced playouts and visit cap oscillation are disabled, tree reuse is enabled.
\item The root uses $c_{\text{FPU}} = 0.2$ just the same as the rest of the search tree instead of $c_{\text{FPU}} = 0.0$. 
\item Resignation is enabled, occurring if both sides agree that for the last 5 turns, the worst MCTS winrate estimate $p$ for the losing side has on each turn been less than 5\%.
\end{itemize}

\pagebreak

\section{Score Maximization} \label{ScoreMaximization}
Unlike most other Go bots learning from self-play, KataGo puts nonzero utility on maximizing (a dynamic monotone function of) the score difference, to improve use for human game analysis and handicap game play.

Letting $x$ be the final score difference of a game, in addition to the utility for winning/losing:
\[ u_{\text{win}}(x) = \text{sign}(x) \in \{-1,1\} \] 
We also define the score utility:
\[ u_{\text{score}}(x) = c_{\text{score}} f\left(\frac{x - x_0}{b}\right) \]
where $c_{\text{score}}$ is a parameter controlling the relative importance of score, $x_0$ is a parameter for centering the utility curve, $b \in [9,19]$ is the width of the board and $f: \mathbb{R} \rightarrow (-1,1)$ is the function:
\[ f(x) = \frac{2}{\pi} \arctan(x) \]

\begin{figure}[h]
\centerline{
\includegraphics[width=3.1in]{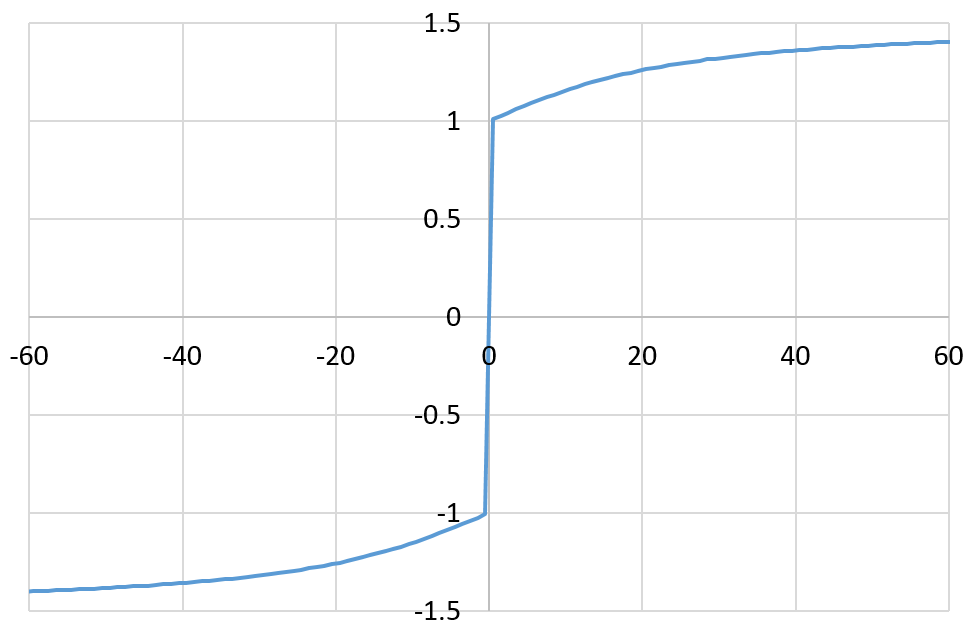}
}
\caption{Total utility as a function of score difference, when $x_0 = 0$ and $b = 19$ and $c_{\text{score}} = 0.5$.}
\label{ScoreUtility}
\end{figure}

At the start of each search, the utility is re-centered by setting $x_0$ to the mean $\hat{\mu}_s$ of the neural net's predicted score distribution at the root node. The search proceeds with the aim to maximize the sum of $u_{\text{win}}$ and $u_{\text{score}}$ instead of only $u_{\text{win}}$. Estimates of $u_{\text{win}}$ are obtained using the game outcome value prediction of the net as usual, and estimates of $u_{\text{score}}$ are obtained by querying the neural net for the mean and variance $\hat{\mu}_s$ and $\hat{\sigma}_s^2$ of its predicted score distribution, and computing:
\[ E(u_{\text{score}}) \approx \int_{-\infty}^{\infty} u_{\text{score}}(x) N(x,\hat{\mu}_s,\hat{\sigma}_s^2) dx \]
where the integral on the right is estimated quickly by interpolation in a precomputed lookup table.

Since similar to a sigmoid $f$ saturates far from $0$, this provides an incentive for improving the score in simple and likely ways near $x_0$ without awarding overly large amounts of expected utility for pursuing unlikely but large gains in score or shying away from unlikely but large losses in score. For KataGo's main run, $c_{\text{score}}$ was initialized to $0.5$, then adjusted $0.4$ after the first two days of training.

\end{appendices}

\end{document}